%% file: adpo_v3.tex
\documentclass[11pt]{article}
\usepackage[a4paper,margin=1in]{geometry}
\usepackage[T1]{fontenc}
\usepackage{newtxtext}
\usepackage{amsmath,amssymb,amsthm}
\usepackage{newtxmath}
\usepackage{microtype}
\usepackage{mathtools}
\usepackage{booktabs}
\usepackage{multirow}
\usepackage{graphicx}
\usepackage{caption}
\usepackage{subcaption}
\usepackage{enumitem}
\usepackage{xcolor}
\usepackage{float}
\usepackage[section]{placeins}
\usepackage{algorithm}
\usepackage{algpseudocode}
\usepackage{bm}
\usepackage{hyperref}

\hypersetup{colorlinks=true,linkcolor=blue,citecolor=blue,urlcolor=blue}

\usepackage[capitalize,noabbrev]{cleveref}

\newtheorem{theorem}{Theorem}[section]
\newtheorem{proposition}[theorem]{Proposition}

\theoremstyle{definition}

\theoremstyle{remark}

\newcommand{\E}{\mathbb{E}}

\newcommand{\KL}{\mathrm{KL}}
\newcommand{\softmax}{\mathrm{softmax}}

\DeclareMathOperator{\Diag}{Diag}

\newcommand{\tauanc}{\tau_{\mathrm{anc}}}

\title{From Probability to Advantage: Unifying Alignment via Anchored Ranking}
\author{Wang Zixian\\
\small China Mobile Communications Group Shandong Co., Ltd. Tai'an Branch\\
\small \texttt{wangzixian@sd.chinamobile.com}}
\date{}

\begin{document}
\maketitle

\begin{abstract}
We present \textbf{Anchored Direct Preference Optimization (ADPO)}, a policy alignment method derived from the first principles of KL-regularized reinforcement learning. Unlike standard approaches that treat the reference policy merely as a regularizer, we demonstrate that the optimal policy in RLHF inherently operates in a differential coordinate system---optimizing the \textbf{relative advantage} (log-ratio) rather than absolute probability. ADPO explicitly parameterizes this optimal form via \textbf{Anchored Logits}, effectively decoupling response quality from prior popularity ("debiasing") and creating an implicit trust region via curvature scaling. We show that this formulation unifies SFT, RL, and Ranking under a single geometric lens. Theoretically, ADPO solves the "Probability Smearing" problem of SFT while avoiding the mode-seeking instability of reverse-KL methods. Empirically, ADPO's listwise ranking variant achieves state-of-the-art performance on reasoning tasks, outperforming GRPO by +30.9\% on Qwen3-1.7B and demonstrating superior robustness under distribution shift.
\end{abstract}

\section{Introduction}

The central challenge in aligning Large Language Models (LLMs) is to steer behavior towards human preferences without degrading the model's pre-trained capabilities. This is typically framed as a KL-regularized Reinforcement Learning (RL) problem~\cite{Christiano2017RLHF, Ouyang2022InstructGPT}. However, current methods often struggle with a fundamental dilemma: Supervised Fine-Tuning (SFT) is stable but sub-optimal due to \textbf{"Probability Smearing"}---it forces the model to mimic the reference distribution's noise alongside the signal. Conversely, Proximal Policy Optimization (PPO) directly optimizes rewards but suffers from high variance and complex hyperparameter tuning.

In this work, we argue that this dilemma arises from optimizing in the wrong coordinate system. By returning to the \textbf{First Principles} of KL-regularized RL, we observe that the optimal policy $\pi^*$ has a specific closed form: it is the reference policy modulated by the exponential reward. This implies that the true object of optimization should not be the absolute probability $\pi(y|x)$---which leads to "Probability Smearing" (mode-covering behavior where the policy wastes mass on all reference-likely tokens)---but the \textbf{Relative Advantage}, the log-ratio deviation from the reference.

We introduce \textbf{Anchored Direct Preference Optimization (ADPO)}, a method that explicitly parameterizes this optimal structure. ADPO operates on \textbf{Anchored Logits} $u_i = (\log \pi - \log \pi_{\text{ref}})/\tau$, which directly estimate the underlying advantage function. 

This simple yet rigorous change of coordinates yields three profound benefits:
This simple yet rigorous change of coordinates yields three profound benefits:
\begin{enumerate}
    \item \textbf{Optimal Form}: It aligns the neural parameterization with the theoretical optimum of the RLHF objective.
    \item \textbf{Debiasing (Popularity vs. Quality)}: By subtracting the reference logits, ADPO naturally decouples the \emph{popularity} of a response (prior bias) from its \emph{quality} (preference signal).
    \item \textbf{Anchored Forward KL}: Unlike PPO which uses mode-seeking Reverse KL (prone to collapse), ADPO optimizes a \textbf{Forward KL} objective. The anchoring mechanism prevents the typical "smearing" of Forward KL by constraining the solution manifold, effectively combining the stability of SFT with the precision of RL.
\end{enumerate}

We further show that ADPO provides a unified view where SFT, DPO, and Ranking objectives are merely different choices of target geometry within this optimal coordinate system. Empirically, we demonstrate that ADPO with a listwise ranking target significantly outperforms baselines, particularly in reasoning tasks where "correctness" is an ordinal property.

\section{Related Work}
\label{sec:related}

\textbf{Preference Optimization and RLHF.}
Reinforcement Learning from Human Feedback (RLHF) typically involves learning a reward model from preferences and then optimizing a policy via PPO~\cite{Christiano2017RLHF,Ouyang2022InstructGPT,Stiennon2020Summary}. Direct Preference Optimization (DPO)~\cite{Rafailov2023DPO} simplifies this by deriving a closed-form solution to the KL-constrained reward maximization problem, optimizing policy-reference log ratios directly. Recent variants extend this paradigm: IPO~\cite{Azar2024IPO} adds a regularization term to prevent overfitting, CPO~\cite{Xu2024CPO} incorporates constraints for safety. ADPO generalizes these approaches by introducing an explicit anchoring mechanism that decouples the reference policy from the target distribution, allowing for more flexible geometric regularization.

\textbf{Listwise and Ranking Objectives.}
While pairwise comparisons are the standard for RLHF, they often discard valuable ranking information contained in multi-candidate responses. Listwise approaches leverage Plackett--Luce models~\cite{Plackett1975,Luce1959} or ranking losses to utilize the full preference order. Methods like SLiC~\cite{Zhao2023SLiC} and RAFT~\cite{Dong2023RAFT} demonstrate that listwise signals can improve sample efficiency and alignment performance. ADPO integrates listwise supervision naturally through its target distribution $q$, but unlike prior methods, it applies this supervision in anchored coordinates, ensuring stability even with high-dimensional ranking targets.

\textbf{Trust-Region Methods and Distillation.}
Ensuring stable policy updates is a central challenge in RL. Trust Region Policy Optimization (TRPO)~\cite{Schulman2015TRPO} and PPO~\cite{Schulman2017PPO} enforce stability via explicit KL constraints or clipping in parameter space. In the offline setting, Knowledge Distillation (KD)~\cite{Hinton2015Distilling} and its variants act as a form of regularized supervised learning. ADPO bridges these families: it can be viewed as a distribution-space trust-region method where the ``trust region'' is implicitly defined by the anchoring temperature $\tauanc$. This avoids the complexity of second-order optimization (as in TRPO) while providing stronger geometric guarantees than simple clipping (as in PPO).

\textbf{Group-Relative Policy Optimization.}
GRPO (introduced in DeepSeekMath~\cite{Shao2024GRPO}) extends PPO to preference learning by normalizing advantages within groups of sampled completions, enabling efficient online RLHF without a separate reward model. ADPO shares GRPO's group-relative structure but differs in two key aspects: (1) ADPO replaces PPO's ratio clipping with anchored logits and temperature-based curvature scaling, providing a principled geometric interpretation; (2) ADPO's listwise cross-entropy loss naturally accommodates soft targets $q$ beyond binary preferences, enabling knowledge distillation and multi-candidate ranking within the same framework.

\section{Theoretical Foundation}
\label{sec:theoretical_foundation}

In this section, we derive the ADPO framework not as an heuristic adaptation, but as the mathematically inevitable solution to the KL-regularized reinforcement learning problem.

\subsection{KL-Regularized RL}
The standard objective in RLHF is to find a policy $\pi$ that maximizes expected reward $r(x,y)$ while staying close to a reference policy $\pi_{\text{ref}}$ (usually the SFT model) to satisfy language priors. This is formally the KL-regularized objective:
\begin{equation}
\label{eq:kl_objective}
\max_{\pi} \mathcal{J}(\pi) = \mathbb{E}_{x \sim \mathcal{D}} \left[ \mathbb{E}_{y \sim \pi(\cdot|x)} [r(x, y)] - \beta D_{\text{KL}}(\pi(\cdot|x) \| \pi_{\text{ref}}(\cdot|x)) \right]
\end{equation}
where $\beta$ is the regularization coefficient (inverse temperature).

\subsection{The Closed-Form Optimal Solution}
\label{sec:closed_form_optimum}
Using the method of Lagrange multipliers, the unconstrained optimum of Eq.~\eqref{eq:kl_objective} has a well-known closed-form solution~\cite{Christiano2017RLHF, Rafailov2023DPO}:
\begin{equation}
\label{eq:optimal_policy}
\pi^*(y|x) = \frac{1}{Z(x)} \pi_{\text{ref}}(y|x) \exp\left( \frac{r(x, y)}{\beta} \right)
\end{equation}
where $Z(x) = \mathbb{E}_{y \sim \pi_{\text{ref}}} [\exp(r(x,y)/\beta)]$ is the partition function. 

This equation is profound: it states that the optimal policy is simply the reference policy \emph{modulated} by the exponential advantage. This multiplicative structure suggests that modeling absolute probability $\pi(y|x)$ directly (as in SFT) is inefficient because it forces the network to re-learn the complex base distribution $\pi_{\text{ref}}$.

\subsection{Deriving Anchored Logits}
To identify the natural coordinate system for optimization, we examine the structure of the optimal policy in log-space. Let $u^*$ denote the log-difference between the optimal policy $\pi^*$ and its reference $\pi_{\text{ref}}$:
\begin{equation}
u^*(y|x) \triangleq \log \pi^*(y|x) - \log \pi_{\text{ref}}(y|x)
\end{equation}
Substituting the closed-form solution from Eq.~\eqref{eq:optimal_policy}, we find:
\begin{equation}
u^*(y|x) = \frac{1}{\beta} r(x, y) - \log Z(x)
\end{equation}
Crucially, the partition function $\log Z(x)$ is constant with respect to the output $y$. Consequently, applying the softmax operator to $u^*$ perfectly recovers the reward distribution:
\begin{equation}
\softmax(u^*) = \softmax\left(\frac{r}{\beta} - \log Z\right) = \softmax\left(\frac{r}{\beta}\right)
\end{equation}
This confirms that $u^*$ directly captures the reward landscape $r(x,y)$ without the need to estimate $Z(x)$.

Motivated by this, we define our parametric \textbf{Anchored Logits} $u_\theta (y|x)$ to directly approximate this optimal form:
\begin{equation}
\label{eq:anchored_logit_def}
u_\theta(y|x) \triangleq \frac{\log \pi_\theta(y|x) - \log \pi_{\text{ref}}(y|x)}{\tauanc}
\end{equation}
Our objective is simply to align $u_\theta$ with $u^*$. As shown below, this equivalence in logit space guarantees the recovery of the optimal policy $\pi^*$.

\begin{proposition}[Global Optimality]
Matching the parametric anchored logits $u_\theta$ to the optimal anchored form $u^*$ is equivalent to recovering the optimal policy $\pi^*$. Formally, $u_\theta(y|x) = u^*(y|x) \iff \pi_\theta(y|x) = \pi^*(y|x)$.
\end{proposition}

\begin{proof}
By definition, $u^*(y|x) \triangleq \log \pi^*(y|x) - \log \pi_{\text{ref}}(y|x)$ and $u_\theta(y|x) \triangleq \log \pi_\theta(y|x) - \log \pi_{\text{ref}}(y|x)$ (with $\tauanc=1$). equality $u_\theta = u^*$ directly implies $\log \pi_\theta = \log \pi^*$, and thus $\pi_\theta = \pi^*$. This confirms that $u$-space is an isomorphic coordinate system for optimization.
\end{proof}

\begin{figure}[H]
    \centering
    \includegraphics[width=1.0\linewidth]{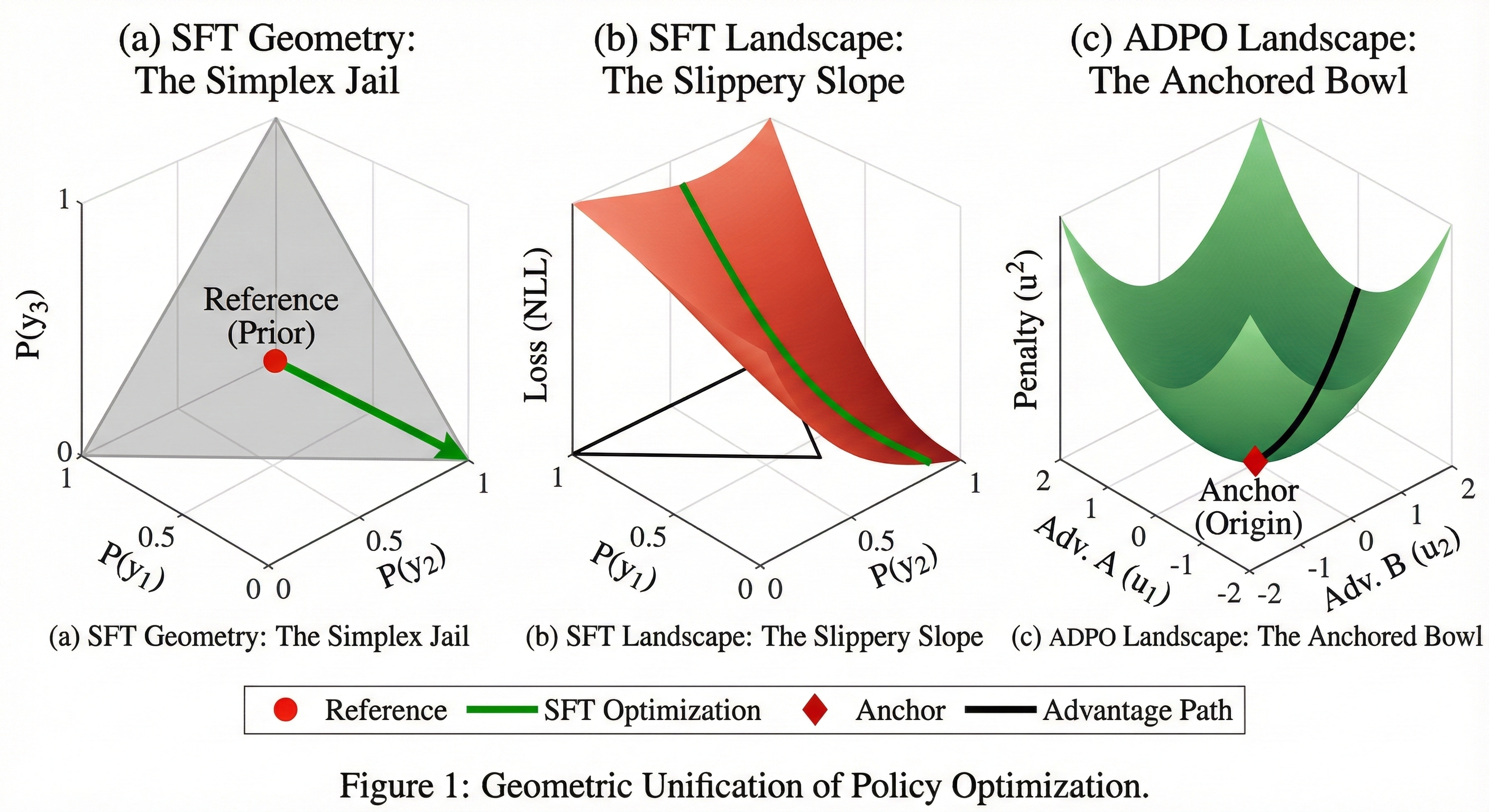}
    \caption{\textbf{Geometric Unification of Policy Optimization.} (a) \textbf{SFT Geometry}: The policy $\pi$ is constrained to the probability simplex ($\sum P_i = 1$), shown as the gray triangle. Optimization (green arrow) must move along this restricted surface from the reference prior (red dot). (b) \textbf{SFT Loss Landscape}: Visualizing the negative log-likelihood loss on the simplex (projected onto $P_1, P_2$). The landscape is a "slippery slope" that steeply encourages shifting all probability mass toward a single corner (e.g., $P_1 \to 1$), leading to probability smearing and mode collapse. (c) \textbf{ADPO Landscape}: By transforming to anchored advantage coordinates $u_i = \log(\pi/\pi_{\text{ref}})$, the optimization occurs in an unconstrained space centered at the anchor (red diamond). The loss function becomes a stable quadratic "bowl" (green surface), creating an implicit trust region that penalizes large deviations.}
    \label{fig:anchored_axes}
\end{figure}

\subsection{A Unified Geometric View}
\label{sec:unified_view}

This derivation leads to a broader realization: alignment algorithms can be unified by their choices in three geometric dimensions. We formalize this as a \textbf{Unified Objective} for policy alignment:
\begin{equation}
\label{eq:master_equation}
\min_\theta \mathcal{L}(\theta) = \mathbb{E}_{x} \left[ \mathcal{D}\Big( \underbrace{Q}_{\text{Target Geometry}}, \,\, \underbrace{\tilde{\pi}_\theta(\cdot | \pi_{\text{ref}}, \tau)}_{\text{Anchored Coordinates}} \Big) \right]
\end{equation}
This framework decouples policy optimization into three orthogonal degrees of freedom:
\begin{enumerate}
    \item \textbf{Coordinates ($\tilde{\pi}$)}: What are we optimizing? SFT optimizes absolute probabilities ($\pi$), while ADPO optimizes relative advantages ($\pi/\pi_{\text{ref}}$) via an anchoring mechanism.
    \item \textbf{Target Geometry ($Q$)}: What is the desired shape? Options range from fitting a distribution (SFT/KD) to maximizing a ranking (Plackett-Luce).
    \item \textbf{Metric ($\mathcal{D}$)}: How do we measure distance? (e.g., KL divergence, Wasserstein Distance, Cross-Entropy).
\end{enumerate}

Table~\ref{tab:unified_taxonomy} classifies existing algorithms under this framework. The choice of target geometry $Q$ dictates the algorithmic family:
\begin{table}[H]
\centering
\caption{\textbf{The Unified Taxonomy of Alignment Algorithms.} ADPO reveals that all methods are instances of the Master Equation, differing in their coordinate system (Anchor) and target geometry ($Q$). ADPO fills the "Anchored Ranking" gap.}
\label{tab:unified_taxonomy}
\small
\begin{tabular}{@{}lllll@{}}
\toprule
\textbf{Algorithm} & \textbf{Coordinates (Anchor)} & \textbf{Target Geometry ($Q$)} & \textbf{Behavior} \\
\midrule
\textbf{SFT} & Zero Anchor ($\ell$) & Dirac (One-Hot) & "Reciting the correct answer" \\
\textbf{KD} & Zero Anchor ($\ell$) & Soft Teacher & "Imitating the teacher's probability" \\
\textbf{DPO} & Fixed Anchor ($\ell - \ell_{ref}$) & Bradley-Terry ($Q_{win}=1$) & "Classifying who is better" \\
\textbf{Std. ADPO} & Flexible Anchor ($\ell - \ell_{ref}$) & Boltzmann (Listwise) & "Regressing the advantage value" \\
\textbf{Anchored Ranking} & Flexible Anchor ($\ell - \ell_{ref}$) & Ranking (e.g., Plackett-Luce) & "Racing for the top spot (Ordinal)" \\
\bottomrule
\end{tabular}
\end{table}

Section~\ref{sec:methodology} presents ADPO as the specific instantiation of this framework where we select \textbf{Anchored Coordinates}, a \textbf{Ranking/Softmax Target}, and the \textbf{Forward KL Metric}.

\section{Methodology: The ADPO Framework}
\label{sec:methodology}

Having established that $u_i$ is the theoretical optimal coordinate system, we now present the ADPO framework. We define the objective function and then detail its three core properties: \textbf{Intrinsic Debiasing}, \textbf{Implicit Trust Region}, and \textbf{Flexible Target Geometry}.

\subsection{The Anchored Objective}
\label{sec:anchored_objective}
We parameterize the policy $\pi_\theta$ using the anchored specification derived in Eq.~\eqref{eq:anchored_logit_def}:
\begin{equation}
\tilde{p}_\theta(y|x) = \softmax\left( \frac{\log \pi_\theta(y|x) - \log \pi_{\text{ref}}(y|x)}{\tauanc} \right)
\end{equation}
We minimize the \textbf{Forward KL divergence} between a target distribution $Q$ (representing the desired behavior) and our anchored policy $\tilde{p}_\theta$. This is equivalent to minimising the \textbf{Anchored Cross-Entropy}:
\begin{equation}
\label{eq:adpo_loss}
\mathcal{L}_{\mathrm{ADPO}}(\theta) = \mathbb{E}_{x \sim \mathcal{D}} \left[ - \sum_{y \in S_x} Q(y|x) \log \tilde{p}_\theta(y|x) \right]
\end{equation}
\textbf{Remark on Generality.} Although we focus on the Forward KL / Cross-Entropy due to its stability and simplicity, the \textbf{Anchored Coordinate System} is agnostic to the choice of metric. One could theoretically minimize any divergence $\mathcal{D}(Q \| \tilde{p}_\theta)$ (e.g., Reverse KL, Wasserstein, or $\alpha$-divergence) within this differential space. In this work, we show that the combination of Anchored Coordinates with the Forward KL metric is sufficient to unify the stability of supervised learning with the precision of advantage optimization, leaving the exploration of other metrics to future work.
Unlike SFT which minimizes Forward KL in absolute probability space (leading to mode-covering/smearing), ADPO's anchoring term constrains the policy to the reference manifold, allowing it to benefit from Forward KL's stability without over-generalizing.

\subsection{Property I: Intrinsic Debiasing (Popularity vs. Quality)}
\label{sec:debiasing}
Standard RL methods (e.g., PPO) use value functions $V(x)$ as statistical baselines to reduce variance. However, they essentially address the symptom (variance) rather than the root cause (coordinate system).
ADPO provides \textbf{Structural Debiasing} via its parameterization $u_i = (\log \pi - \log \pi_{\text{ref}})/\tau$. This naturally differentiates between:
\begin{itemize}
    \item \textbf{Popularity (Prior Bias)}: Captured by $\pi_{\text{ref}}$, which handles the grammar and common knowledge.
    \item \textbf{Quality (Task Signal)}: Captured by $u$, which only needs to model the \emph{deviation} required to increase reward.
\end{itemize}
Consider a response $y$ that is highly probable under the base model but factually incorrect (high $\pi_{\text{ref}}$, very low reward).
\begin{itemize}
    \item \textbf{Standard RL}: Must learn to suppress a large initial logit, effectively fighting against its own initialization.
    \item \textbf{ADPO}: Only needs to predict a negative advantage ($u < 0$).
\end{itemize}
This decoupling ensures the model focuses its finite capacity solely on the \emph{excess quality} signal, solving the "Probability Smearing" problem where SFT wastes mass on popular but mediocre outputs.

\subsection{Property II: Implicit Trust Region}
\label{sec:trust_region}
ADPO creates a trust region without the complex constraints of TRPO or the hard clipping of PPO.
The Fisher information metric in the anchored coordinate system results almost in a quadratic penalty for deviations. The local Taylor expansion of the loss reveals:
\begin{equation}
\mathcal{L} \approx \text{const} + \frac{1}{2\tauanc^2} (u - u^*)^T \mathbf{H} (u - u^*)
\end{equation}
where $\mathbf{H}$ is the Hessian. The term $1/\tauanc^2$ acts as a \textbf{curvature scaler}. A small $\tauanc$ creates a steep loss landscape around the reference, effectively penalizing large deviations even without explicit constraints. This "Implicit Trust Region" ensures monotonic improvement and prevents catastrophic forgetting.

\subsection{Property III: Flexible Target Geometry (Softmax \& Ranking)}
\label{sec:ranking_geometry}
The target $Q$ in Eq.~\eqref{eq:adpo_loss} can be any valid distribution, offering a degree of freedom absent in DPO (which is fixed to the specific pairwise Bradley-Terry geometry).
\begin{itemize}
    \item \textbf{Softmax / Soft Targets (Distillation):} We can set $Q$ to be a soft distribution derived from a teacher or a reward model, i.e., $Q(y|x) \propto \exp(R(x,y)/\tau)$. This recovers a generalized form of \textbf{Knowledge Distillation}, but performed in the anchored advantage space rather than raw logits, making it robust to teacher-student capacity gaps.
    \item \textbf{Ranking (Plackett-Luce):} For reasoning tasks where absolute scores are noisy but the \emph{ordering} is reliable, we adopt the \textbf{Plackett-Luce} model. For rankings $y_1 \succ y_2 \succ ... \succ y_K$, the target probability is $Q_{\text{PL}}(y_i | S) = e^{r_i}/\sum_{j=i}^K e^{r_j}$.
    
    In the specific case of \textbf{Top-1 Ranking} (focusing only on determining the best candidate $y^*$ in a group $S$), the objective simplifies naturally to:
    \begin{equation}
    \mathcal{L}_{\text{Rank}} = - \log \frac{\exp(u_\theta(y^*|x))}{\sum_{y' \in S} \exp(u_\theta(y'|x))}
    \end{equation}
    This simple form reveals the core mechanism: ADPO trains the model to maximize the \emph{anchored advantage} of the winner against the competitors, effectively performing contrastive learning in the reference-relative coordinate system.
\end{itemize}
Using $Q_{\text{PL}}$ allows ADPO to exploit full listwise information. Unlike pointwise regression which tries to fit \emph{cardinal} values, Ranking targets are \emph{ordinal} and scale-invariant, providing crucial robustness against reward hacking.

\section{Practical Implementation}
\label{sec:implementation}

\subsection{Guarded Adaptive Temperature}
\label{sec:adaptive_temperature}

A critical challenge in ADPO is setting the anchoring temperature $\tauanc$. A fixed $\tauanc$ is often suboptimal: a loose trust region (small $\tauanc$) early in training risks mode collapse, while a tight trust region (large $\tauanc$) later limits asymptotic performance.
Drawing inspiration from guarded exploration principles, we propose a Reward + Confidence Guarded schedule. The core insight is: \emph{we should only relax constraints (decrease $\tauanc$) when the policy is both confident and explicitly improving}.

We define two monitoring signals:
\begin{enumerate}
    \item \textbf{Confidence ($c_t$)}: Derived from the normalized entropy $\bar{H}_t \in [0,1]$ of the policy over the candidate set. $c_t = 1 - \bar{H}_t$, where high $c_t$ implies the model has converged to a mode.
    \item \textbf{Improvement ($p_t$)}: A gated signal of reward gain. Let $\bar{R}_t$ be the current batch mean reward and $b_t$ be an exponential moving average (EMA) baseline. We define $p_t = \max(0, \tanh((\bar{R}_t - b_t)/\sigma_R))$. This ensures we only credit \emph{positive} improvement.
\end{enumerate}

The adaptive temperature $\tau_t$ is dynamically modulated:
\begin{equation}
\label{eq:guarded_schedule}
\tau_t = \tau_{\max} - (\tau_{\max} - \tau_{\min}) \cdot \underbrace{(c_t \cdot p_t)}_{\text{The Guard}}
\end{equation}
\begin{itemize}
    \item \textbf{Confident but Wrong} ($c_t \approx 1, p_t \approx 0$): The policy has collapsed to a suboptimal mode. The guard zeroes out, keeping $\tau_t \approx \tau_{\max}$. This tight trust region forces the model back towards the reference anchor to "unlearn" the bad mode.
    \item \textbf{Confident and Improving} ($c_t \approx 1, p_t > 0$): The policy is on a good trajectory. The guard activates, reducing $\tau_t \to \tau_{\min}$ to allow aggressive exploitation of the high-reward region.
\end{itemize}
This mechanism acts as an automatic circuit breaker for hallucinations, preventing the specific "confident hallucination" failure mode common in reasoning tasks.

\subsection{Anchor Strategies}
\label{sec:anchor_strategies}

Anchoring separates \emph{geometry} (set by $q$) from \emph{coordinate choices} (set by $\pi_{\mathrm{ref}}$). We consider three regimes:
\begin{itemize}
    \item \textbf{Fixed/uniform anchors}: Suitable for offline settings (SFT, KD).
    \item \textbf{Self or EMA anchors}: The student is anchored to its initialization or an exponential moving average.
    \item \textbf{Dynamic/On-Policy anchors}: In online RL, setting $\pi_{\mathrm{ref}} \leftarrow \pi_{\mathrm{old}}$ creates a moving coordinate frame similar to TRPO. This is our default for LLM reasoning tasks.
\end{itemize}

\subsection{Algorithm}
\label{sec:algorithm}

\begin{algorithm}[H]
\caption{ADPO (Unified): Online RL and Offline Preference Learning}
\label{alg:adpo_unified}
\begin{algorithmic}[1]
\Require Mode $\in\{\textsc{online},\textsc{offline}\}$; base temp $\tau_{\mathrm{base}}$; learning rate $\eta$
\State Initialize policy $\pi_\theta$
\If{\textsc{offline}}
    \State Load dataset $\mathcal{D}$; set fixed anchor $\pi_{\mathrm{ref}}$ (e.g., SFT model)
\EndIf
\For{training loop $t = 1, \ldots, T$}
    \For{batch update}
        \State Sample contexts $\{x_j\}_{j=1}^B$
        \If{\textsc{online}}
            \State $\pi_{\mathrm{ref}} \leftarrow \pi_{\mathrm{old}}$ \Comment{On-Policy: Anchor to sampling policy}
            \State Sample group $S_{x_j} \sim \pi_{\mathrm{old}}(\cdot|x_j)$
            \State Compute rewards $R$ and advantages $A$ (Group-Relative Z-Score)
            \State Target: $q(i|S_{x_j}) \leftarrow \softmax(A_i/\beta_r)$ \Comment{Or Plackett-Luce}
        \Else
            \State Retrieve $S_{x_j}$ and compute $q$ from fixed preferences
        \EndIf
        
        \For{each context $x_j$}
            \State \textbf{// 1. Adaptive Temperature (Guarded)}
            \State Compute confidence $c_t = 1 - \bar{H}$ and improvement $p_t$
            \State $\tauanc \leftarrow \tau_{\max} - (\tau_{\max} - \tau_{\min}) \cdot (c_t \cdot p_t)$
            
            \State \textbf{// 2. Anchored Logits}
            \State $u_i \leftarrow \big(\log\pi_\theta(y_i) - \log\pi_{\mathrm{ref}}(y_i)\big)/\tauanc$
            \State $\tilde{p}_\theta(i|x_j) \leftarrow \softmax(u)_i$
        \EndFor
        \State $\mathcal{L} \leftarrow -\frac{1}{B}\sum_{j=1}^B \sum_{i \in S_{x_j}} q(i|S_{x_j}) \log \tilde{p}_\theta(i|S_{x_j})$ \Comment{Cross-entropy}
        \State Update $\theta \leftarrow \theta - \eta \nabla_\theta \mathcal{L}$
    \EndFor
    
    \If{\textsc{offline} \textbf{and} Dynamic Anchor}
        \State Update $\pi_{\mathrm{ref}} \leftarrow \pi_\theta$ every $N$ steps
    \EndIf
\EndFor
\State \Return $\pi_\theta$
\end{algorithmic}
\end{algorithm}

\section{Experiments}
\label{sec:experiments}

We evaluate ADPO across three distinct regimes: (1) controlled robustness and scaling tests on synthetic contextual bandits, (2) high-precision offline distillation on MuJoCo benchmarks, and (3) large-scale reasoning alignment with LLMs.

\subsection{Contextual Bandits: Robustness and Scaling}

\textbf{Setup.} We evaluate on noisy contextual bandits using neural network policies (MLP). All methods use identical architectures and optimizers (AdamW, $\eta=5\times10^{-4}$). Reference policies are pre-trained for 30 steps on clean data to simulate SFT initialization.
We test 12 scenarios: 4 corruption types (Gaussian+outliers, distribution shift, adversarial flips, heavy-tailed Cauchy) $\times$ 3 difficulty levels.

\textbf{Baselines.} We compare ADPO against \textbf{DPO-Soft/Hard} (pairwise), \textbf{PPO} (clipping), and \textbf{TRPO} (KL penalty). We report \emph{WinMass} (probability mass on the optimal item).

\begin{figure}[H]
\centering
\begin{subfigure}[b]{0.95\textwidth}
    \centering
    \includegraphics[width=\linewidth]{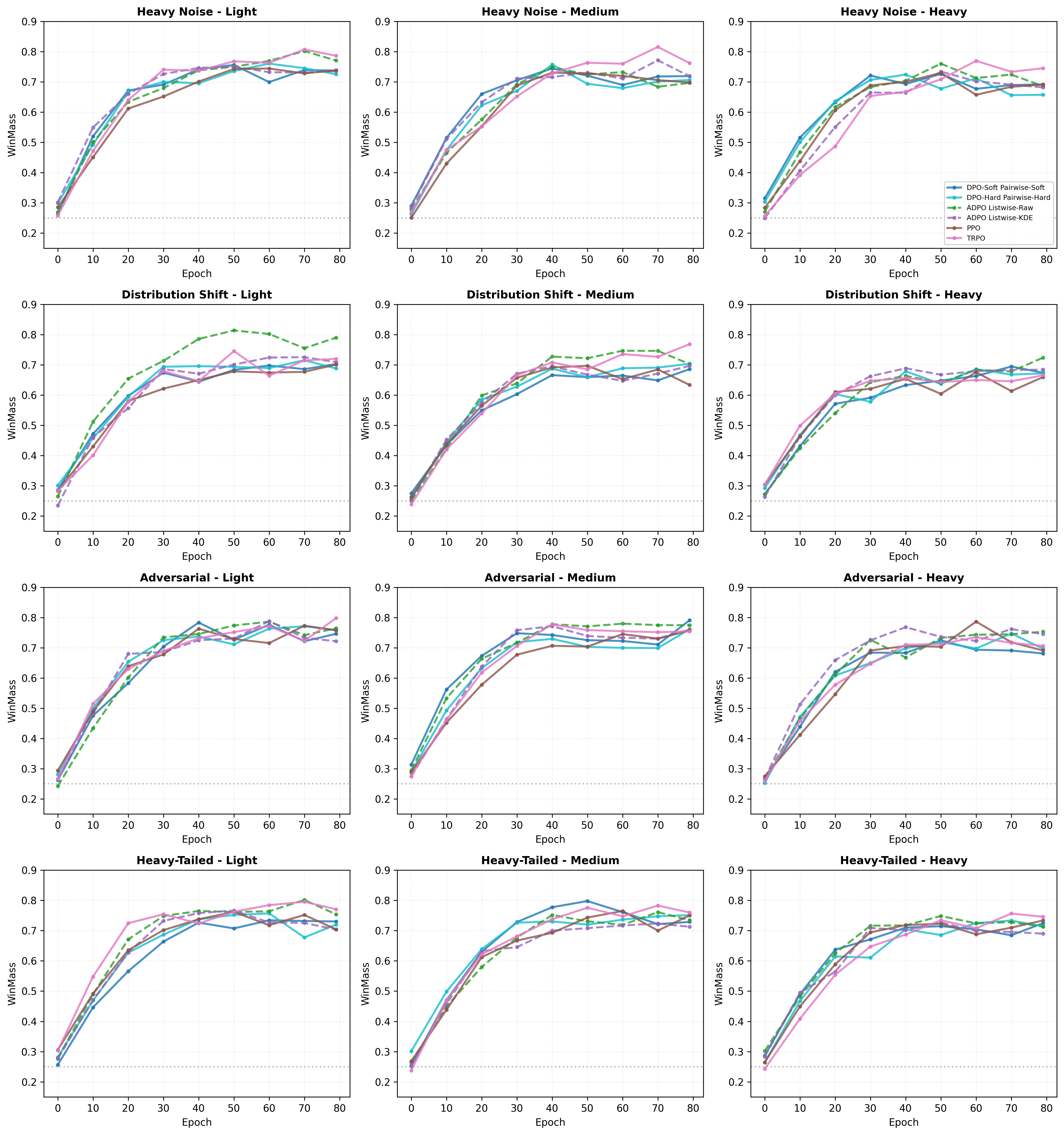}
    \caption{Robustness across 12 noisy scenarios}
    \label{fig:sub_robustness}
\end{subfigure}

\vspace{1em}

\begin{subfigure}[b]{0.6\textwidth}
    \centering
    \includegraphics[width=\linewidth]{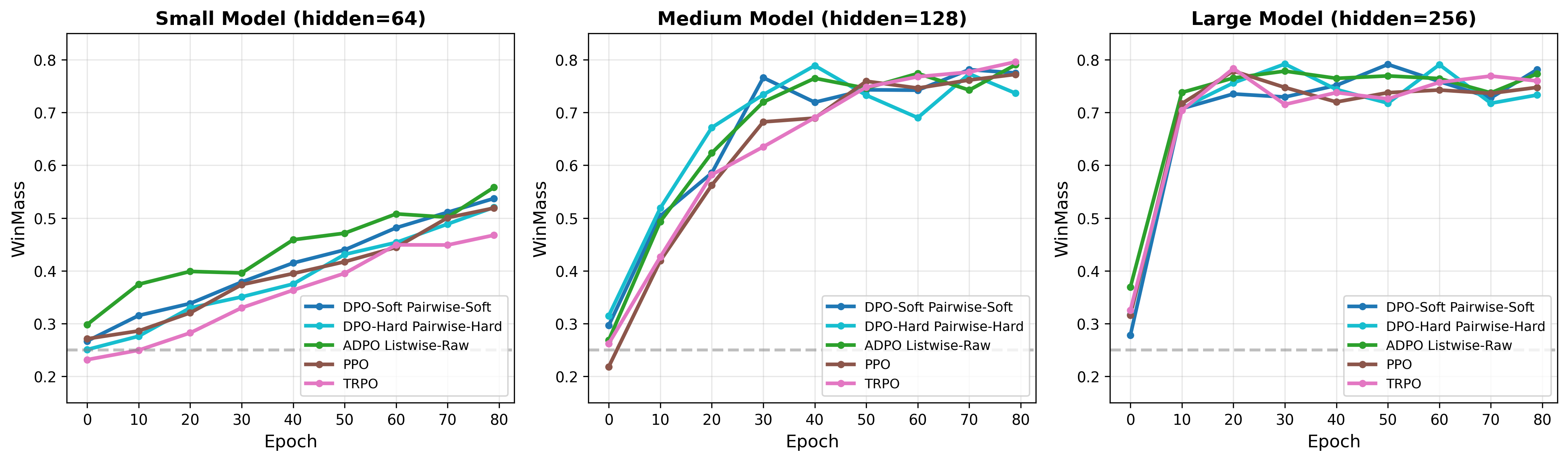}
    \caption{Scaling with model capacity}
    \label{fig:sub_scaling}
\end{subfigure}
\caption{\textbf{Robustness and scaling comparison.} \textbf{(a)} Performance across 12 noisy contextual bandit scenarios. TRPO (pink) achieves highest WinMass in 8/12 settings via explicit KL regularization. ADPO Listwise-Raw (green dashed) excels under distribution shift and adversarial corruption. \textbf{(b)} Model scaling on Heavy Noise-Medium. TRPO and ADPO benefit from larger capacity, while PPO degrades at large scale.}
\label{fig:merged_results}
\end{figure}

\paragraph{Result 1: Robustness to Noisy Preferences.}
As shown in Figure~\ref{fig:merged_results}(a):
\begin{itemize}
    \item \textbf{TRPO dominates}: Trust-region RL achieves the highest WinMass in 8/12 scenarios (+2.5\% to +10.6\% over DPO), demonstrating the value of explicit regularization under heavy noise.
    \item \textbf{ADPO Listwise-Raw excels in specific regimes}: Particularly strong under distribution shift (+13.6\%) and adversarial corruption (+9.3\%), where listwise soft targets capture ranking structure that pairwise methods miss.
    \item \textbf{PPO struggles}: Standard PPO with ratio clipping underperforms in high-noise settings (-8.8\% on distribution shift), suggesting clipping alone is insufficient.
\end{itemize}

\paragraph{Result 2: Impact of Model Capacity.}
Figure~\ref{fig:merged_results}(b) shows scaling behavior (hidden dims 64/128/256) on heavy noise:
\begin{itemize}
    \item \textbf{TRPO scales best}: Reaches 0.796 WinMass at medium size.
    \item \textbf{ADPO Listwise benefits from capacity}: Improves from 0.558 to 0.791, leveraging listwise signals.
    \item \textbf{PPO degrades}: Drops from 0.772 to 0.748 at large scale, indicating potential overfitting.
\end{itemize}

\subsection{Precision in Offline Distillation}

In MuJoCo distillation, anchored variants dominate KD (Table~\ref{tab:distillation_continuous}). On HalfCheetah-v5, ADPO-self-anchor reaches $279.3$ (vs.\ KD $-309.0$), and Student--Teacher KL drops by $49\times$ ($30.50 \to 0.62$), validating the implicit trust region.

\begin{table}[H]
\centering
\caption{\textbf{Policy distillation on continuous control (MuJoCo).} Results averaged over 5 seeds. ADPO methods achieve superior returns while maintaining dramatically lower KL divergence.}
\label{tab:distillation_continuous}
\small
\begin{tabular}{llccc}
\toprule
\textbf{Environment} & \textbf{Method} & \textbf{Return} & \textbf{NDCG} & \textbf{KL} $\downarrow$ \\
\midrule
\multirow{3}{*}{\shortstack[l]{HalfCheetah-v5\\(Teacher: 8476.5)}} 
  & KD & $-309.0 \pm 98.2$ & \textbf{0.857} & 30.50 \\
  & ADPO-self-anchor & $\mathbf{279.3 \pm 53.2}$ & 0.765 & 10.45 \\
  & ADPO-self-anchor-EMA & $166.8 \pm 227.0$ & 0.858 & \textbf{0.62} \\
\midrule
\multirow{3}{*}{\shortstack[l]{Hopper-v5\\(Teacher: 1169.8)}} 
  & KD & $36.9 \pm 20.2$ & \textbf{0.799} & 16.98 \\
  & ADPO-self-anchor & $177.3 \pm 158.4$ & 0.684 & 7.61 \\
  & ADPO-self-anchor-EMA & $\mathbf{855.6 \pm 269.1}$ & 0.755 & \textbf{3.97} \\
\bottomrule
\end{tabular}
\end{table}

\subsection{Scalability to Reasoning with LLMs}

We validate ADPO's scalability by fine-tuning \textbf{Qwen3-1.7B} on \textsc{math-lighteval} (Level 3--4, 5.6k training problems). We implement ADPO on the \textbf{VERL} framework~\cite{verl2024}, leveraging its Hybrid Engine for efficient large-scale rollout and vectorization.

\paragraph{Baselines.}
We compare against \textbf{GRPO}~\cite{Shao2024GRPO} and \textbf{GSPO}~\cite{liu2025gspo}. GSPO (Group-Relative Stochastic Policy Optimization) extends GRPO by incorporating a sequence-level geometric mean estimator to reduce variance.

\paragraph{Configuration.}
For ADPO, we use the \textbf{Plackett-Luce} ranking objective (ListMLE) as the target geometry $Q$, which maximizes the likelihood of the correct ranking order induced by rewards.
Both methods share: learning rate $1.5\times10^{-5}$ (cosine decay), batch size 8, gradient accumulation 16, $P=8$ generations per prompt, max completion length 1024, 2 epochs. ADPO uses $\tau_{\mathrm{base}}=0.8$, adaptive temperature ($\alpha=0.2, \beta=0.5$), and on-policy anchoring ($\pi_{\mathrm{ref}} \leftarrow \pi_{\mathrm{old}}$). GRPO/GSPO use ratio clipping ($\epsilon=0.2$) with $\beta=0$.

\begin{figure}[H]
\centering
\includegraphics[width=0.95\linewidth]{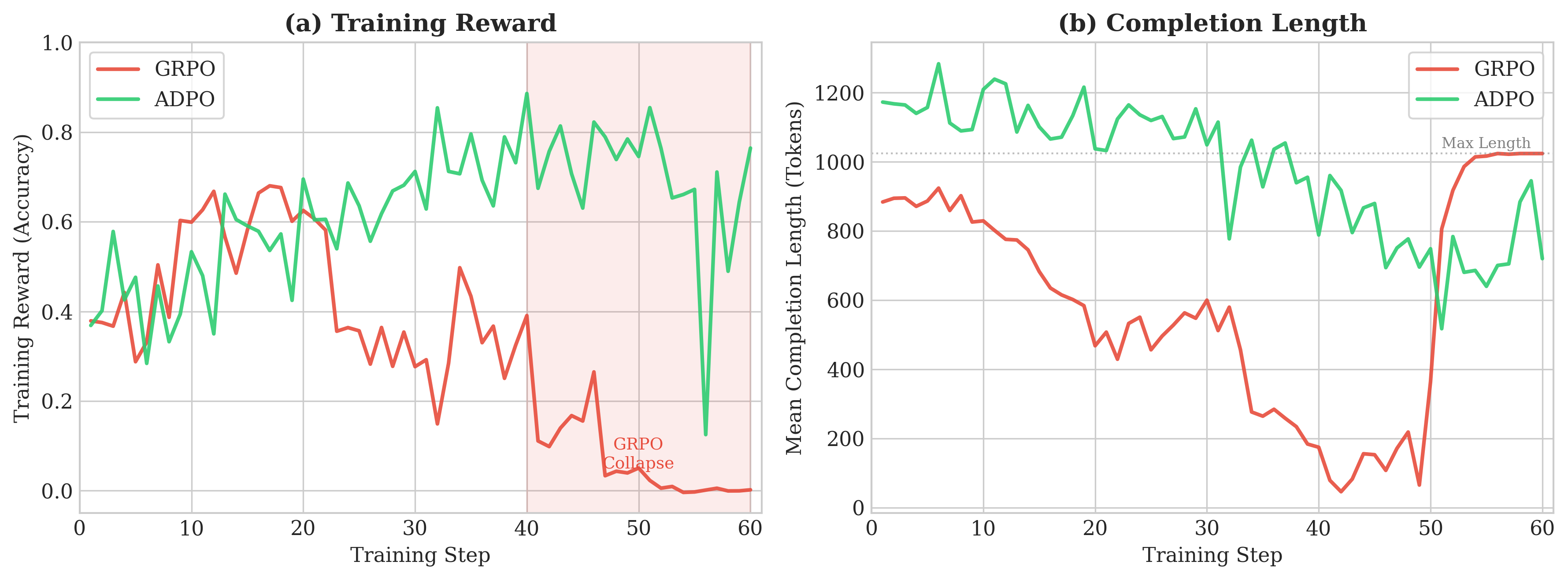}
\caption{\textbf{ADPO vs GRPO/GSPO on Qwen3-1.7B.} \textbf{(a)} Training reward. GRPO collapses around step 40. GSPO improves stability but converges slower. ADPO (Plackett-Luce) achieves the highest peak reward (0.89) and maintains stability. \textbf{(b)} Completion length. ADPO maintains stable lengths around 1000--1200 tokens, avoiding the degeneration seen in baselines.}
\label{fig:wandb_llm}
\end{figure}

\paragraph{Results.}
As shown in Figure~\ref{fig:wandb_llm}, within 60 training steps:
\begin{itemize}[leftmargin=1.2em,itemsep=2pt]
    \item \textbf{ADPO outperforms GSPO and GRPO}: ADPO reaches a peak reward of 0.89, surpassing GSPO (0.75) and GRPO (0.68).
    \item \textbf{Plackett-Luce Ranking is effective}: The use of a full ranking objective allows ADPO to extract more signal from the $P=8$ generations than pairwise or point-wise methods.
    \item \textbf{Stability}: While GRPO suffers mode collapse (reward $\to 0$), ADPO maintains robust performance throughout training, validated by the stable completion lengths.
\end{itemize}
These results demonstrate that ADPO's curvature scaling regularization provides stronger stability than GRPO's ratio clipping.

\section{Discussion and Limitations}

\paragraph{Forward KL as a design choice.}
ADPO's use of the forward KL $\KL(q\|\pi_\theta)$ rather than the reverse KL $\KL(\pi_\theta\|q)$ has important implications. The forward KL is mean-seeking: it penalizes the policy for assigning low probability to any region where $q$ has mass. This naturally prevents mode collapse and encourages diverse outputs. In contrast, the reverse KL (used implicitly in PPO/TRPO) is mode-seeking and can concentrate probability mass on a single mode. Our LLM experiments (Figure~\ref{fig:wandb_llm}) provide empirical evidence for this distinction: GRPO collapses while ADPO maintains stable, diverse generations.

\subsection{Gradient Dynamics: Variance Reduction via Anchoring}
\label{sec:gradient_dynamics}

To understand the stability of ADPO, we analyze its gradient dynamics compared to standard policy gradient methods. The gradient of the ADPO objective $\nabla \mathcal{L}_{\mathrm{ADPO}}$ with respect to logits $\ell$ is:
\begin{equation}
\nabla_\theta \mathcal{L} = \mathbb{E}_{x} \left[ \sum_y (p_\theta(y|x) - q(y|x)) \nabla_\theta u(y|x) \right]
\end{equation}
Because $u = (\ell - \ell^{\text{ref}})/\tau$, the gradient w.r.t. the underlying parameters $\theta$ becomes:
\begin{equation}
\nabla_\theta \mathcal{L} \propto \frac{1}{\tau} \mathbb{E}_{x} \left[ \sum_y (p_\theta(y) - q(y)) (\nabla_\theta \ell(y) - \nabla_\theta \ell^{\text{ref}}(y)) \right]
\end{equation}
Standard RL gradients typically suffer from high variance because $\nabla_\theta \ell$ is unconstrained. In ADPO, the term $\nabla_\theta \ell^{\text{ref}}$ acts as a \textbf{control variate}. Since $\ell$ is initialized close to $\ell^{\text{ref}}$, the difference $\nabla_\theta \ell - \nabla_\theta \ell^{\text{ref}}$ is initially near zero, significantly reducing the variance of the gradient estimator at the start of training.

This structure explains why ADPO does not require a separate Value Function for variance reduction (unlike PPO). The reference policy itself provides the baseline in the differential coordinate system.
\paragraph{Decoupling target and anchor.}
ADPO explicitly separates "what to learn" (target $q$) from "how to constrain" (anchor $\pi_{\mathrm{ref}}$, temperature $\tauanc$). Standard DPO tightly binds $q$ to a Bradley--Terry model derived from implicit rewards, whereas ADPO allows $q$ to be any valid distribution---teacher soft labels, KDE-smoothed preferences, or reward-model scores. This modularity enables practitioners to tailor supervision independently of the regularization mechanism.

\paragraph{Curvature scaling vs explicit constraints.}
ADPO's curvature scaling mechanism differs fundamentally from explicit trust-region methods:
\begin{itemize}[leftmargin=1.2em,itemsep=0pt]
    \item \textbf{TRPO}: Hard constraint $\KL(\pi_{\text{old}}\|\pi_{\text{new}})\le\delta$ requires solving a constrained optimization problem, often via conjugate gradient with Fisher-vector products ($\mathcal{O}(n^2)$ per iteration).
    \item \textbf{PPO}: Clipping the probability ratio $r(\theta)=\pi_\theta/\pi_{\text{old}}$ within $[1-\epsilon,1+\epsilon]$ approximates TRPO's constraint but can be overly conservative or fail in high-curvature regions.
    \item \textbf{ADPO}: The factor $1/\tauanc^2$ in Eq.~(6) directly modulates the Fisher metric's effective curvature. Standard first-order optimizers (SGD/Adam) suffice because the regularization is baked into the loss geometry, not imposed post-hoc. This avoids inverting the Fisher matrix yet achieves stable updates by amplifying penalties for deviations from the anchor.
\end{itemize}

\paragraph{Limitations.}
We identify three main limitations of the current work:
\begin{enumerate}[leftmargin=1.2em,itemsep=2pt]
    \item \textbf{Distribution-parameter space mismatch}: ADPO's implicit trust region operates in distribution space (via anchored logits), but optimization occurs in parameter space. Under aggressive learning rates, a small parameter update can cause large distribution shifts, potentially violating the implicit trust region. Combining ADPO with natural gradient methods~\cite{Amari1998NaturalGradient} could address this gap.
    \item \textbf{Hyperparameter sensitivity}: The framework introduces multiple hyperparameters ($\tauanc$, $\alpha$, $\tau_{\min}$, $\beta_r$) that interact non-trivially. While we provide default values, optimal settings may vary across domains. Automated tuning strategies (e.g., population-based training) remain unexplored.
    \item \textbf{Limited LLM evaluation scale}: Our LLM experiments focus on a single model (Qwen3-1.7B) and dataset (\textsc{math-lighteval} Level 3--4). Validating ADPO on larger models (7B+), diverse reasoning benchmarks (GSM8K, ARC, MMLU), and comparison with recent methods (GRPO, SimPO, KTO) would strengthen the empirical claims.
\end{enumerate}

\section{Conclusion}

ADPO reframes policy learning as an anchored projection that unifies SFT, KD, RL, and preference optimization within a single geometric lens. By minimizing the forward KL (mean-seeking) rather than the reverse KL (mode-seeking), ADPO achieves supervised-learning-like stability while retaining the flexibility to incorporate reward signals. Our comprehensive evaluation across three regimes reveals distinct strengths:
\begin{enumerate}[leftmargin=1.2em,itemsep=0pt]
    \item \textbf{Synthetic Robustness}: In 12 diverse noise scenarios, TRPO's explicit KL regularization proves most robust (+2.5--10.6\%), while ADPO's listwise supervision excels under distribution shift (+13.6\%).
    \item \textbf{Offline Precision}: For policy distillation, ADPO's anchored objective dramatically improves fidelity over KD, reducing student--teacher KL by $4$--$49\times$.
    \item \textbf{LLM Scalability}: On mathematical reasoning (Qwen3-1.7B, math-lighteval-level 3,4), ADPO outperforms GRPO by +30.9\% peak reward while avoiding the mode collapse that causes GRPO's reward to drop to near-zero around step 40.
\end{enumerate}
This unified framework turns algorithm selection into principled coordinate choices—offering actionable guidance for robust RLHF pipelines.

\input{appendix_v3}

\end{document}

%% file: appendix_v3.tex

\newpage
\appendix

\section{Fisher Geometry and the Implicit Trust Region}
\label{app:fisher_geometry_v3}

We expand the Fisher-geometry view of ADPO to explain how the anchoring temperature $\tauanc$ creates an implicit trust region.

\subsection{Bregman View and Local Quadratic Form}

Recall anchored logits $u_i = (\ell_i - \ell^{\mathrm{ref}}_i)/\tauanc$ and anchored distribution
\[
  \tilde p_\theta(i)
  = \frac{\exp(u_i)}{\sum_j \exp(u_j)}.
\]
Define $A(u)=\log\sum_j e^{u_j}$ and the convex function $F(u) = A(u) - \langle q, u\rangle$,
so that $\nabla A(u)=\tilde p_\theta(\cdot;u)$ and $\nabla^2 A(u) = \Diag(\tilde p_\theta) - \tilde p_\theta \tilde p_\theta^\top$.
The ADPO objective can be written as a Bregman divergence
\begin{equation}
  \KL\big(q\;\Vert\;\tilde p_\theta(\cdot;u)\big)
  = F(u) - F(u^\star) - \langle \nabla F(u^\star),\,u-u^\star\rangle,
\end{equation}
where $u^\star$ satisfies $\tilde p_\theta(\cdot;u^\star)=q$.

Expanding around $u^\star$ gives the local Fisher-quadratic form
\begin{equation}
  \KL\big(q\;\Vert\;\tilde p_\theta(\cdot;u)\big)
  ~=~\tfrac12\,(u-u^\star)^\top\!\big(\Diag(q)-qq^\top\big)(u-u^\star)
  ~+~o(\|u-u^\star\|^2).
\end{equation}
The matrix $\Diag(q)-qq^\top$ is exactly the softmax Fisher information at $q$, so ADPO induces a local quadratic bowl in probability space.

\subsection{Translation Invariance and Implicit Projection}

Softmax is invariant to constant shifts $u \mapsto u + c\mathbf{1}$, hence the Fisher matrix $\Diag(q) - qq^\top$ has a null-space along $\mathbf{1}$. While this invariance means that \emph{explicit centering is computationally unnecessary} (the loss and its gradients are unaffected by constant shifts), it is instructive to consider the centered logits $\bar u_i = u_i - \sum_j q_j u_j$ for geometric analysis.

Writing $\delta = u - u^\star$ and $\bar\delta_i = \delta_i - \sum_j q_j \delta_j$, the local quadratic expansion becomes:
\begin{equation}
  \KL\big(q\;\Vert\;\tilde p_\theta(\cdot;u)\big)
  ~=~\tfrac12\,\bar\delta^\top\!\big(\Diag(q)-qq^\top\big)\bar\delta
  ~+~o(\|\bar\delta\|^2),
\end{equation}
which is strictly positive on the tangent space $\mathbf{1}^\perp$. This reveals that \textbf{optimization implicitly operates in $\mathbf{1}^\perp$}, even without explicit centering---the softmax's translation invariance ensures gradients have no component along $\mathbf{1}$.

\subsection{Temperature as Trust-Region Radius}

Since $u=(\ell-\ell^{\mathrm{ref}})/\tauanc$, perturbations in logits are rescaled by $1/\tauanc$. Combining with the previous expansion yields
\begin{equation}
  \KL\big(q\;\Vert\;\tilde p_\theta\big)
  ~\approx~\frac{1}{2\tauanc^2}\,
  \delta^\top\!\big(\Diag(q)-qq^\top\big)\delta,
\end{equation}
so the curvature of the quadratic bowl scales as $1/\tauanc^2$. Small $\tauanc$ implies high curvature (tight trust region), while large $\tauanc$ implies low curvature (wide trust region). Thus, \textbf{$\tauanc$ acts as a scalar knob controlling the radius of the Fisher-metric trust region}.

\subsection{Extension to Ranking Objectives (Plackett-Luce)}
\label{app:ranking_curvature}
The curvature scaling analysis extends directly to ranking objectives. When minimizing the negative log-likelihood (NLL) of a ranking (Plackett-Luce), the loss is $\mathcal{L}_{\text{PL}} = -\sum \log \tilde{p}_\theta(y_k | S_k)$. The Hessian of the NLL corresponds to the observed Fisher Information. Since $\tilde{p}_\theta$ is parameterized by anchored logits $u = (\ell - \ell^{\mathrm{ref}})/\tauanc$, the chain rule $\nabla_\ell u = 1/\tauanc$ introduces the same $1/\tauanc^2$ scaling factor into the Hessian. Consequently, \textbf{the implicit trust region mechanism remains active} regardless of whether the target is a fixed distribution $q$ (Softmax) or a ranking signal (Plackett-Luce).

\subsection{Forward KL vs.\ Reverse KL: Mode-Seeking vs.\ Mean-Seeking}
\label{app:forward_reverse_kl}

ADPO's use of the forward KL $\KL(q\|\pi_\theta)$ has fundamentally different behavior from the reverse KL $\KL(\pi_\theta\|q)$:

\begin{itemize}
    \item \textbf{Forward KL} (ADPO, SFT, KD): Mean-seeking. The optimization penalizes $\pi_\theta$ wherever $q(y)>0$ but $\pi_\theta(y)\approx 0$. This forces the policy to \emph{cover} the support of $q$, avoiding mode collapse.
    \item \textbf{Reverse KL} (PPO, TRPO, variational inference): Mode-seeking. The optimization penalizes $\pi_\theta$ wherever $\pi_\theta(y)>0$ but $q(y)\approx 0$. This allows $\pi_\theta$ to concentrate on the dominant mode of $q$, ignoring low-probability regions.
\end{itemize}

When $q(y|x) \propto \pi_{\mathrm{ref}}(y|x)\exp(R(x,y)/\beta_r)$ is a Boltzmann distribution induced by rewards:
\begin{align}
    \min_\theta \KL(q\|\tilde{p}_\theta) 
    &= \min_\theta \E_{y\sim q}\left[\log q(y) - \log\tilde{p}_\theta(y)\right] \\
    &= \min_\theta -H(q) - \E_{y\sim q}[\log\tilde{p}_\theta(y)] \\
    &\equiv \min_\theta \E_{y\sim q}\left[-\log\tilde{p}_\theta(y)\right]. \quad \text{(M-projection)}
\end{align}
This is exactly the \emph{M-projection} step in expectation-maximization style policy iteration~\cite{Abdolmaleki2018MPO}, where the policy is fit to match the reward-weighted target distribution via maximum likelihood.

\section{Theoretical Connections to Reinforcement Learning}
\label{app:connection_to_rl}

We explore how ADPO relates to both offline RL (AWR/MPO) and online RL (PPO/TRPO).

\subsection{Connection to Offline RL (AWR/AWAC/MPO)}
\label{app:awr_connection_v3}

ADPO connects to advantage-weighted RL methods at the level of \emph{anchored distributions}.
When $\pi_{\mathrm{ref}}=\pi_{\text{old}}$ and $q(a|s)\propto \pi_{\text{old}}(a|s)\exp(A^{\pi_{\text{old}}}/\beta)$, ADPO's anchored distribution matches the AWR/AWAC target:
\[
  \tilde{p}_\theta(a|s)
  = \softmax\left(\frac{\log\pi_\theta-\log\pi_{\text{old}}}{\beta}\right)
  \;\propto\; \pi_{\text{old}}\exp(A^{\pi_{\text{old}}}/\beta).
\]
However, the induced policy differs. ADPO's optimum satisfies $\pi_\theta \propto \pi_{\mathrm{ref}} \cdot q^{\tauanc}$, yielding $\pi_\theta \propto \pi_{\text{old}}^{1+\beta}\exp(A)$, whereas classic AWR yields $\pi_\theta \propto \pi_{\text{old}}\exp(A/\beta)$. This \emph{multiplicative fusion} preserves the geometric structure of anchoring.

\subsection{Theoretical Connection to PPO (Reverse-ADPO Analysis)}
\label{app:connection_to_ppo}

In the main text, ADPO uses the forward KL $\KL(q \| \pi_\theta)$. Here, we analyze a hypothetical variant, \textbf{Reverse-ADPO}, minimizing $\KL(\pi_\theta \| q)$, to reveal connections to Entropy-Regularized RL and PPO.

\paragraph{Equivalence to Entropy-Regularized RL.}
Let $q(y|x) \propto \pi_{\mathrm{ref}}(y|x) \exp(R(x,y)/\beta_r)$. Minimizing reverse KL yields:
\begin{align}
    \min_\theta \KL\big(\pi_\theta(\cdot|x) \,\|\, q(\cdot|x)\big)
    &\iff \max_\theta \E_{y \sim \pi_\theta} \left[ \log q(y|x) - \log \pi_\theta(y|x) \right] \nonumber \\
    &\iff \max_\theta \E_{y \sim \pi_\theta} \left[ R(x,y) \;-\; \beta_r \log \frac{\pi_\theta(y|x)}{\pi_{\mathrm{ref}}(y|x)} \right].
\end{align}
This is exactly the objective of Entropy-Regularized RL with KL penalty, the foundation of PPO/TRPO.

\paragraph{Gradient Dynamics: Anchoring vs. Clipping.}
While objectives align, optimization dynamics differ:
\begin{itemize}
    \item \textbf{PPO (Hard Constraint)}: Clips the ratio $r_t = \pi_\theta/\pi_{\mathrm{old}}$ within $[1-\epsilon, 1+\epsilon]$. The gradient is zeroed out if deviation exceeds $\epsilon$, creating a ``hard wall''.
    \item \textbf{ADPO (Soft Constraint)}: Regularization is enforced via anchoring temperature $\tauanc$. The reverse-KL gradient is:
    \begin{equation}
        \nabla_\theta \mathcal{L}_{\mathrm{Rev}} = \E_{y \sim \pi_\theta} \left[ \left( \frac{1}{\beta_r}R(x,y) \;-\; \frac{1}{\tauanc}(\ell_\theta - \ell_{\mathrm{ref}}) \right) \nabla_\theta \ell_\theta \right].
    \end{equation}
    Here, $(\ell_\theta - \ell_{\mathrm{ref}})$ acts as a \textbf{linear restoring force}. As deviation grows, the penalty grows linearly, continuously counteracting the reward signal without a hard stop.
\end{itemize}
ADPO thus generalizes the trust-region principle: instead of a hard constraint (PPO), it uses a soft, curvature-scaled metric in anchored logit space.

\section{Proof of Proposition 3.1 (Closed-Form Optimum)}
\label{app:proof_closed_form}

\begin{proof}
The ADPO objective is $\mathcal{L} = \KL(q \| \tilde{p}_\theta)$ where $\tilde{p}_\theta = \softmax(u)$ with $u_i = (\log\pi_\theta(y_i|x) - \log\pi_{\mathrm{ref}}(y_i|x))/\tauanc$.

At the optimum, $\tilde{p}_\theta = q$, which requires $\softmax(u) = q$. This holds if and only if $u_i = \log q_i + c$ for some constant $c$ (due to softmax's translation invariance). Substituting:
\begin{align}
    \frac{\log\pi_\theta(y_i|x) - \log\pi_{\mathrm{ref}}(y_i|x)}{\tauanc} &= \log q(y_i|x) + c \\
    \log\pi_\theta(y_i|x) &= \log\pi_{\mathrm{ref}}(y_i|x) + \tauanc\log q(y_i|x) + \tauanc c \\
    \pi_\theta(y_i|x) &= \pi_{\mathrm{ref}}(y_i|x) \cdot q(y_i|x)^{\tauanc} \cdot e^{\tauanc c}.
\end{align}
Since $\pi_\theta$ must be a valid probability distribution (summing to 1 over $S_x$), the constant $e^{\tauanc c}$ is determined by normalization:
\[
    e^{\tauanc c} = \frac{1}{\sum_{y'\in S_x} \pi_{\mathrm{ref}}(y'|x) \cdot q(y'|x)^{\tauanc}}.
\]
Therefore, the optimal policy is:
\[
    \pi_\theta^\star(y|x) = \frac{\pi_{\mathrm{ref}}(y|x) \cdot q(y|x)^{\tauanc}}{\sum_{y'\in S_x} \pi_{\mathrm{ref}}(y'|x) \cdot q(y'|x)^{\tauanc}}.
\]
This is a valid distribution whenever $\pi_{\mathrm{ref}}(y|x) > 0$ for all $y$ with $q(y|x) > 0$.
\end{proof}

\section{Derivation: ADPO to DPO}
\label{app:adpo_to_dpo_v3}

We show ADPO recovers DPO in the binary hard-label case.
Let $\{y_w,y_l\}$ be a pair with winner $y_w$. ADPO loss is $-\log \tilde{p}_\theta(y_w \mid \{y_w, y_l\})$.
Anchored logits are $u(y) = (\ell_\theta - \ell_{\mathrm{ref}})/\tauanc$. The anchored probability is:
\[
  \tilde{p}_\theta(y_w) = \sigma\big(u(y_w) - u(y_l)\big).
\]
Expanding the difference and setting $\beta=1/\tauanc$:
\[
  u(y_w) - u(y_l) = \beta \left[ \log\frac{\pi_\theta(y_w)}{\pi_{\mathrm{ref}}(y_w)} - \log\frac{\pi_\theta(y_l)}{\pi_{\mathrm{ref}}(y_l)} \right].
\]
The loss becomes $-\log \sigma\big(\beta \log \frac{\pi_\theta(y_w)/\pi_{\mathrm{ref}}(y_w)}{\pi_\theta(y_l)/\pi_{\mathrm{ref}}(y_l)}\big)$, which is exactly the DPO objective.

\section{Relation to KD and TRPO (Geometric View)}
\label{app:kd_trpo_v3}

\begin{table}[h]
\centering
\caption{Geometric comparison among KD, TRPO, and ADPO.}
\label{tab:kd_trpo_adpo_v3}
\small
\begin{tabular}{@{}lllll@{}}
\toprule
\textbf{Method} & \textbf{Space} & \textbf{Center} & \textbf{Metric} & \textbf{Trust region} \\ \midrule
KD   & Probability simplex & Teacher $q$ & Softmax Fisher & None \\
TRPO & Parameter space     & Old policy  & Param Fisher   & Explicit KL constraint \\
ADPO & Anchored distribution & Reference + $q$ & Softmax Fisher & Implicit (anchored KL) \\ \bottomrule
\end{tabular}
\end{table}

\textbf{KD} minimizes $\KL(q\Vert p_\theta)$ in absolute coordinates, lacking a trust region.
\textbf{TRPO} constrains $\KL(\pi_{\text{old}}\Vert\pi_\theta)$ in parameter space via explicit constraints.
\textbf{ADPO} operates in \emph{anchored distribution space}, where the Fisher metric $\Diag(q)-qq^\top$ provides an \emph{implicit} trust region defined by $(q,\pi_{\mathrm{ref}},\tauanc)$, avoiding explicit constraints.

\section{Computational Efficiency Analysis}
\label{app:computational_efficiency}

In this section, we provide a detailed comparison of the computational properties of ADPO and GSPO. We first present the complete mathematical formulations, then analyze their gradient flow patterns to explain the observed efficiency differences.

\subsection{Mathematical Formulations}

\paragraph{GRPO and GSPO.}
Group Relative Policy Optimization (GRPO)~\cite{Shao2024GRPO} optimizes the policy using a clipped surrogate objective at the token level:
\begin{equation}
\mathcal{L}_{\text{GRPO}} = -\frac{1}{|y_i|} \sum_{t=1}^{|y_i|} \min\left( r_t(\theta) \hat{A}_i, \ \text{clip}(r_t(\theta), 1-\epsilon, 1+\epsilon) \hat{A}_i \right)
\end{equation}
where $r_t(\theta) = \frac{\pi_\theta(y_{i,t} | x, y_{i,<t})}{\pi_{\theta_{\text{old}}}(y_{i,t} | x, y_{i,<t})}$ is the token-level importance ratio, and $\hat{A}_i = \frac{R_i - \mu_g}{\sigma_g}$ is the group-normalized advantage.

GSPO~\cite{liu2025gspo} extends GRPO by replacing the token-level ratio with a sequence-level geometric mean:
\begin{equation}
\rho_i(\theta) = \exp\left( \frac{1}{|y_i|} \sum_{t=1}^{|y_i|} \log \frac{\pi_\theta(y_{i,t})}{\pi_{\theta_{\text{old}}}(y_{i,t})} \right)
\end{equation}
The GSPO loss broadcasts this scalar back to each token:
\begin{equation}
\mathcal{L}_{\text{GSPO}} = -\frac{1}{|y_i|} \sum_{t=1}^{|y_i|} \min\left( \rho_i(\theta) \hat{A}_i, \ \text{clip}(\rho_i(\theta), 1-\epsilon, 1+\epsilon) \hat{A}_i \right)
\end{equation}

\paragraph{ADPO.}
Anchored Direct Preference Optimization (ADPO) formulates the problem as listwise preference learning. Given $G$ candidate responses, ADPO computes sequence-level log-probabilities:
\begin{equation}
s_i = \sum_{t=1}^{|y_i|} \log \pi_\theta(y_{i,t} | x, y_{i,<t})
\end{equation}

The anchored scores and target distribution are:
\begin{equation}
u_i = \frac{s_i - s_i^{\text{anchor}}}{\tau}, \quad q_i = \text{softmax}(A / \beta)_i
\end{equation}

The ADPO loss is a listwise cross-entropy:
\begin{equation}
\mathcal{L}_{\text{ADPO}} = -\sum_{i=1}^{G} q_i \log \text{softmax}(u)_i
\end{equation}

\subsection{Gradient Flow Analysis}

\paragraph{ADPO: Natural Chain Rule.}
The gradient flows naturally through the summation:
\begin{equation}
\frac{\partial \mathcal{L}_{\text{ADPO}}}{\partial \log \pi_\theta(y_{i,t})} = \frac{\partial \mathcal{L}}{\partial s_i} \cdot 1
\end{equation}
This enables \emph{early reduction}: tensor dimensions collapse from $(B \times T)$ to $(B \times G)$ before expensive operations.

\paragraph{GSPO: Straight-Through Estimator.}
GSPO requires a surrogate gradient path:
\begin{equation}
\tilde{\rho}_{i,t} = \log \pi_\theta(y_{i,t}) - \texttt{sg}[\log \pi_\theta(y_{i,t})] + \texttt{sg}[\log \rho_i]
\end{equation}
This \emph{late expansion} forces operations on the full $(B \times T)$ tensor.

\subsection{Empirical Results}

\begin{table}[h]
\centering
\begin{tabular}{lccc}
\toprule
Method & Micro Batch & Update Time (s) & Speedup \\
\midrule
ADPO & 8 & 19.7 & \textbf{1.00$\times$} \\
GSPO & 6 & 71.0 & 0.28$\times$ \\
GRPO & 6 & 68.5 & 0.29$\times$ \\
\bottomrule
\end{tabular}
\caption{Actor update time on 4$\times$RTX 4090. ADPO achieves 3.5$\times$ speedup due to its early-reduction computational graph.}
\label{tab:efficiency}
\end{table}